\documentclass[11pt]{article}

\usepackage[final]{acl}

\usepackage{times}
\usepackage{latexsym}
\usepackage{amsmath}
\usepackage{hyperref}
\usepackage{booktabs}
\usepackage{xcolor}
\usepackage{tipa}

\usepackage[T1]{fontenc}

\usepackage[utf8]{inputenc}

\usepackage{microtype}

\usepackage{inconsolata}

\usepackage{graphicx}

\usepackage{pgfplots}

\pgfplotsset{compat=1.17}

\usepackage{amssymb} 
\usepackage{bbold}
\usepackage{adjustbox}
\usepackage{multirow}
\newcommand{\dn}{{\texttt{D$\times$N}}}
\usepackage{array}
\newcolumntype{L}[1]{>{\raggedright\arraybackslash}p{#1}}

\title{Measuring Form and Function in Language Models}

 \author{H\'ector Javier V\'azquez Mart\'inez \and Charles Yang\\
   \texttt{\{hjvm, charles\}@upenn.edu}  \\
   University of Pennsylvania \\
   Department of Linguistics and Computer and Information Science\\
}

\begin{document}
\maketitle

\begin{abstract}
We introduce quantitative metrics from child language research to evaluate language models. Our focus is on the formal syntactic and functional discourse properties of determiners in English, which young children acquire early and accurately. 
We propose \textit{Contextual Alternative Choice} (CAC), a new prompting method which provides targeted tests for both syntactic and discourse knowledge of language.  The  method enables direct comparison of language models against children, and more importantly,   
against independently established benchmarks in empirical research.
No current model trained on a comparable amount of data simultaneously meet both benchmarks like human children, but some very large models do. 
We present our results as  methodological and technical contributions,  with specific emphasis on the cognitive status of language models.\footnote{Our evaluation code and results are available at \\ \url{https://github.com/hjvm/llm-form-and-function}.} 
\end{abstract}

\section{Introduction: LLMs and child language}

Central to American structuralist linguistics \cite{Harris1951, Chomsky1955} is the methodological commitment to {\it distributionalism}: Grammar can be defined formally from a corpus of text (or phonemes) without reference to semantic content. As articulated in {\it Syntactic Structure} \cite[][p50]{Chomsky1957}, the most important goal of linguistic theory under the distributional approach is a {\it discovery procedure}: ''the theory must provide a practical and mechanical method for actually constructing the grammar, given a corpus of utterances.'' In light of the  remarkable success of LLMs, which are primarily trained on text, a set of questions naturally arises: Do LLMs constitute a discovery procedure? Do they encode linguistic knowledge in ways that are comparable to human language users?  
To what extent can they be regarded as a computational model of language acquisition by children? The BabyLM project tackles such fundamental problems in the science of language and cognition, which may in turn provide insight on building data-efficient models to serve  wider linguistic communities \cite{choshenBabyLMTurns42026}.

\begin{figure}[t]
\centering
\includegraphics[width=\columnwidth]{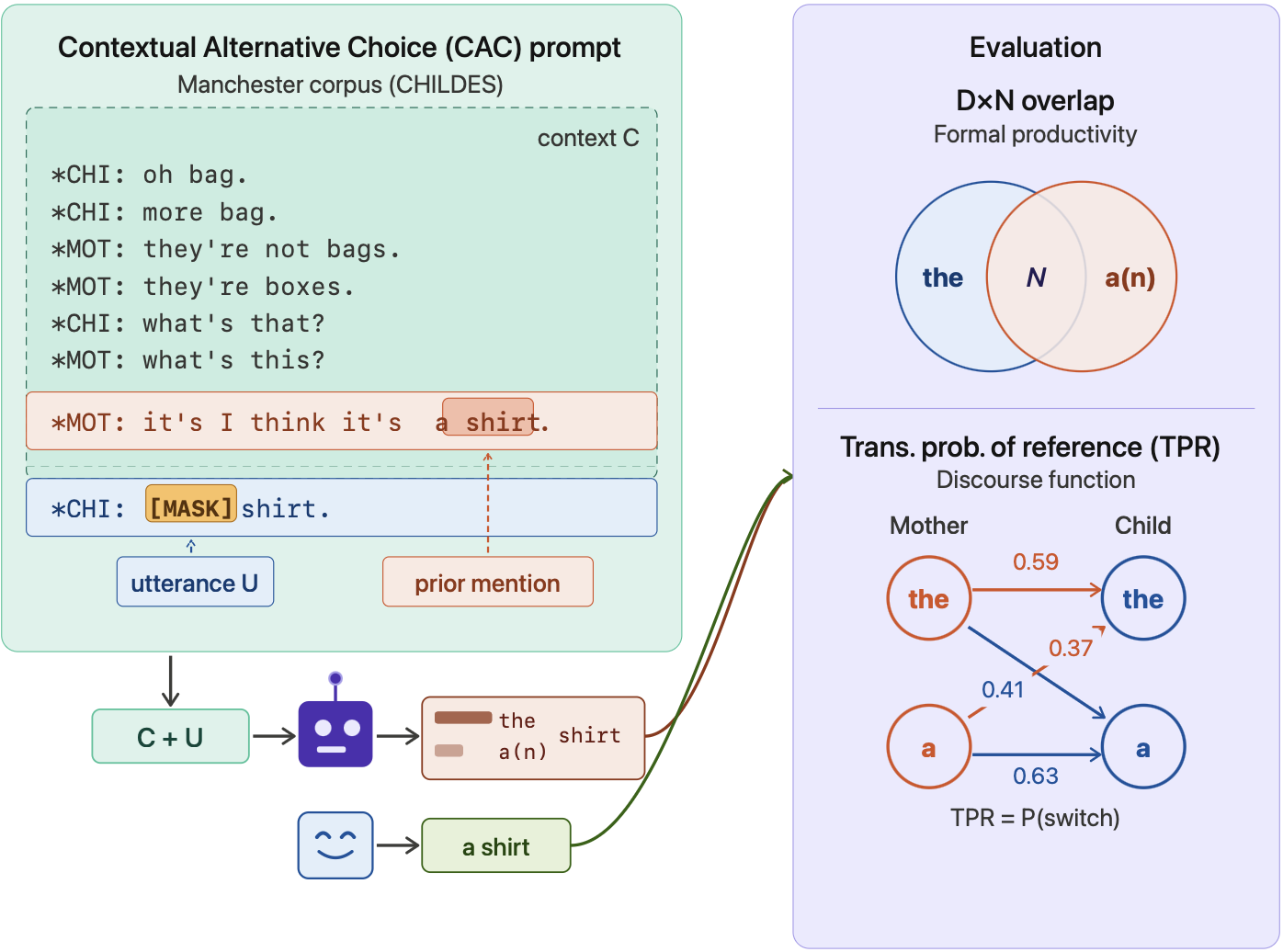}
\caption{Contextual Alternative Choice (CAC) prompting: A target determiner-noun (
\texttt{D}$\times$\texttt{N}) site is identified in caretaker-child dialogue with {\it the} and {\it a} as grammatical choices.  The model is presented with the full preceding discourse context $C$ and predicts the determiner in target utterance U. The resulting determiner preferences are evaluated against two independently motivated benchmarks: formal 
 productivity (overlap) and discourse-functional appropriateness (Transitional Probability of Reference (TPR) described in Section \ref{sec:det-review}.}
\label{fig:cac-diagram}
\end{figure}

To properly approach  these questions requires a framework of reference in which child and LLM language learning can be rigorously compared. To do so, we suggest, calls for closer connection with child language research than present. Our starting point is two striking yet underappreciated parallels between language learning by children and LLMs.  

First, for very different reasons, the inner  mechanisms of human and LLM learning are both poorly understood. There is, quite obviously, very little room for drastic experimentation with human learning.  At the same time, LLMs have so far resisted a complete theoretical understanding despite invasive manipulation of input and model architecture. As such, their external behavior\,---\,what children say and what LLMs do\,---\,remains the primary source of data for assessing their linguistic capacities. Second, the investigation of children and LLMs both face a data poverty problem. The BabyLM challenge is an exception but frontier LLMs are generally trained with closed-source data. Similarly, while a reasonable amount of child-directed data exists in the public domain \cite{CHILDES}, it is amassed from many learners, with no comprehensive record for any particular child. 

In this paper, we exploit these parallels between child and LLM learning by introducing a new set of investigative tools from recent developments in language acquisition. Our proposal is both methodological and empirical, which we take up in turn.

\subsection{Classification vs. generation}
To date, the most popular evaluation  for LLMs,  including the majority in the BabyLM challenge, is based on classification of minimal pairs: grammatical vs. ungrammatical sentences, plausible vs. implausible answers in conceptual understanding and reasoning, correct vs. incorrect pairings between images and captions, etc. Classification datasets, which are easy to scale, are naturally suited for LLMs. However, there are  well-known difficulties relating classification tasks to human linguistic behavior and knowledge. For example, 
the syntactic templates in the widely adopted BLIMP benchmark  frequently introduces semantic and pragmatic anomalies \cite{BLIMP}. However, the influence of these factors on acceptability judgment does not appear simply additive to  syntax, even for binary  tasks \cite{Sprouse2018},  making  it difficult to relate LLM performance to human behavior \cite{ vazquezmartinezEvaluatingNeuralLanguage2023}. 

The classification task is even more problematic in the cognitive setting of BabyLM: Quite simply, there is no ground truth from child language to compare LLMs against. It has long been familiar to researchers that young children cannot provide reliable metalinguistic judgments \cite{DeVilliers1974, 
Thornton2021}. Indeed, most methods readily applicable to adults cannot extend to children. For example, the celebrated Wug test \citep{Berko1958}, the most direct test for linguistic productivity ({\it wug-wugs}), can only reliablely elicit correct responses from first graders; the plural rule, however, is already learned by age 2 
\cite{brownFirstLanguageEarly1973}. The  gap is due to the heavy metalinguistic demands  
 required for learning and using novel words in lab.

For these reasons, child language research typically uses naturalistic production data to benchmark language development. For example, the  size of children's vocabulary,  the simplest component of language, is estimated by only including words that children produce rather than what they (apparently) understand \cite{Fenson1994}. 
Similarly, the  major landmarks of morphological and syntactic acquisition are all established on children's naturalistic production, ranging from the number of morphemes and words per utterance (e.g., mean length of utterance, ``one/two word stage'') to over-generalization errors (e.g., {\it foots}, {\it goed}, {\it I died you}) that mark the productivity of rules  \cite{Ervin1964, Bowerman1982}. The gold standard for assessment is  Brown's 90\% criterion (\citeyear{brownFirstLanguageEarly1973}), which captures both the formal and functional aspects of language:  Acquisition is deem successful if children produce adult-appropriate output in at least 90\% of obligatory contexts.   Originally established on child English, the criterion has been widely adopted in cross-linguistic research \citep{Slobin1985}.

\subsection{Beyond well-formedness}
In recent years, however, Brown's criterion has been under scrutiny. As argued by theorists such as \citet{Tomasello2000}, children's adult-like  usage may simply result from  memorizing, and retrieving, chunks of adult input of varying sizes than a systematic grammar. This should be a familiar refrain in the study of language modeling from $n$-grams to LLMs: It is difficult to know whether certain linguistic capacity can be attributed to generalizations from, or regurgitation of the training data itself \cite{mccoyHowMuchLanguage2023,merrillEvaluatingNGramNovelty2024}. 

In this paper, we introduce two recent benchmarks for measuring child language designed to overcome the deficiencies in Brown's 90\% criterion. Together, they  recognize that well-formedness alone may be too weak of a test for linguistic knowledge. These benchmarks, reviewed in Section \ref{sec:det-review}, compare the statistical profile of a corpus, not against the statistical profile of the input, but against the {\it expected} statistical profile of the target {\it grammar}. As such, they are applicable to any language sample produced by children and LLMs alike, without needing to access input/training data. Our empirical study focuses on English determiners, a richly researched domain that embodies both formal (syntax) and functional (discourse) properties of child language. In order make direct comparisons between LLMs and children, 
we draw inspiration from the familiar cloze task \citep{Taylor1953} from psycholinguistics, especially its use at the discourse level \cite{Chihara1977} and introduce a new LLM prompting method to make targeted grammatical choices in identical contexts as children (Section \ref{sec:cac}). 
We use this framework to  evaluate determiner use by two-year-olds and also by 45 language models (Section \ref{sec:results}) in the BabyLM Project. 
No model trained on a comparable amount of input data meets both benchmarks although two models trained on very large amount of data do. In Section \ref{sec:related-work} and \ref{sec:conclusion}, we discuss related work and the implication of our work in the broader context of LLM research, with specific emphasis on their status as cognitive models of language.

\section{The statistics of determiners}
\label{sec:det-review}

Determiners provide an excellent window into language acquisition. As functional words, determiners' role in language is largely formal, lacking the referential salience and semantic concreteness of content words (nouns, verbs, adjectives, etc.). Nevertheless, determiners are identified as a formal category in infancy and play an important role to establish  content word categories such as nouns; see \citet{Shi2024} for review. Determiners also appear early and frequently in child language production and have proven a fertile ground of research.

\subsection{Measuring determiner productivity}
\label{subsec:det-overlap}
Children's determiner usage is essentially error-free from the get-go \cite{Valian1986}: they must have learned an adult-like grammar by Brown's 90\% criterion. This conclusion has been challenged in recent years \cite{Tomasello2000}. Adult-like usage may result from  memorization and retrieval of input, which would be lexically specific rather than genuinely productive. For example, the English determiners {\it the} and {\it a}\footnote{Following the practice in child language, the phonological variant {\it an} is merged with {\it a} in the present study.} are productive because they can interchangeably combine with any singular noun given suitable context.
A widely adopted numerical measure, {\it overlap}, has been used to quantify  determiner-noun (henceforth {\dn}) productivity \cite{Pine1996}. Given a corpus, overlap is the proportion of noun types that appear with both {\it the} and {\it a} out of those appearing with either {\it the} or {\it a}.  The overlap value in child language corpora is usually quite low (20-40\%), and considerably lower than their caretakers, leading to claims that children do not have a productive determiner grammar but only a collection of lexically specific collocations  from the input.

The overlap measure has generated a large body of quantitative tests for children's grammar but almost all rely on comparisons with, or statistical analysis of, caregiver data to determine the extent of memorization;  see Section \ref{sec:related-work} for details. As such, these methods are inherently limited by the available input data, which at best is only a tiny fraction of the totality. 
By contrast, a formal statistical test  \citep{yangOntogenyPhylogenyLanguage2013} overcomes this obstacle. It derives the expected overlap value generated by a productive grammar, i.e., one that combines {\dn} independently. 
The test builds on two key statistical properties of language, one universal and the other specific to \texttt{D$\times$N} in English. First, the frequencies of nouns are assumed to follow \citet{zipfHumanBehaviorPrinciple1949}'s Law. As such, if a corpus of $S$ \texttt{D$\times$N} combinations involves $N$ unique nouns,  the expected frequency of the noun of rank $r$ is $S/(rH_{N})$ where $H_{N} = \sum_{i=1}^{N} 1/i$. Second, in \texttt{D$\times$N} combinations, nouns tend to have a ``favorite'' (more frequent) determiner. For example, {\it bathroom} greatly favors {\it the} over {\it a} but for {\it bath}, the reverse is true.  This imbalance, referred to as {\it bias} ($b$), is quantified in the aggregate as follows:
\begin{equation}
    b = \frac{\sum\limits_{n \in N} \max(C_{\text{the} \times n}, \,\,C_{\text{a}\times n})}{\sum\limits_{n \in N} (C_{\text{the} \times n}+ C_{\text{a} \times n})}
    \label{eq:bias}
\end{equation}
\noindent where $C_{\text{the/a}\times n}$ is the frequency of {\it the}/{\it a} combined  with $n$ in the corpus. Children do not need to learn these frequencies: they presumably do not  track personal hygiene ({\it a bath}) or  bodily functions ({\it the bathroom}). Rather, the bias value reflects the vagaries of life  in language use, 
and is thus remarkably stable in the aggregate: 0.82 in the billion-word COCA corpus \cite{COCA}, almost identical to the values in the child and caretaker speech analyzed in Section \ref{sec:results}.

These two statistical properties greatly enhance the applicability of the productivity test. For a corpus, one only $S$, $N$, and $b$ are need to calculate the expected value of overlap; see Appendix \ref{app:dxn-derivation} for details. 
If a  comparison between expected and empirical overlap values from a corpus does not show a  significant difference, one can conclude that the corpus is consistent with a productive grammar.

\subsection{Measuring determiner reference}
As emphasized in Section 1, linguistic knowledge is more than well-formedness in classification tests. To fully acquire  determiners, it is not sufficient to know they freely combine with nouns: their communicative functions are equally important.  The determiners {\it a} and {\it the} differ in their referential properties (definiteness, specificity, etc.) in context \cite{Lyons1999}; these functional roles have also been studied in language acquisition.  Three-year-olds can interpret determiner reference in context with high accuracy (85\%; \citealt{Maratsos1976}).  Analyzing children's naturalistic production,  \citet{Rozendaal2008} conclude that two-year-olds already show adult-like determiner usage with respect to specificity, definiteness and givenness. However, these authors also note that interpreting children's determiner usage requires a reconstruction of the child's mental model of discourse, which is both subjective and laborious.

\citet{GleitmanYang2022} propose  a statistical measure of determiner function that eliminates  interpretative guesswork. They focus on contexts where one speaker uses a determiner in response to a determiner used by their interlocutor when discussing the same topic noun: In general, only one of two options is acceptable.   The following two snippets of conversation are instances where the child follows the caretaker with the same determiner; changing it would be highly unnatural albeit formally grammatical. 

\fbox{Mother}: Is it {\bf the dog} or the little boy?

\fbox{Child}: {\bf The dog}  won't stand up properly. 

\fbox{Mother}: I'll make you {\bf a gate}.

\fbox{Child}: Found one. Found {\bf a gate}.

By contrast, the following are instances of determiner change: retaining the same determiner is infelicitous.

\fbox{Mother}: Have you ever had {\bf an itch}?

\fbox{Child}: That's {\bf the itch} over there.

\fbox{Mother}: What happens in {\bf the carwash}, John?

\fbox{Child}: Train's had {\bf a carwash}. 

\citet{GleitmanYang2022} introduce a statistical measure of determiner discourse function:  The {\it transitional probability of reference} (TPR) is the probability with which a speaker {\it changes} determiner following interlocutor while holding the topic noun constant, which can be represented as state transitions in Fig \ref{fig:cac-diagram}. For example, in the four transitions above, TPR is 0.5.  Children's TPR when talking to caretakers can be compared against adults' TPR when talking to each other, which serves a baseline benchmark. A baseline is motivated by the fact that the frequency of communicative needs (e.g., retaining or shifting reference for definiteness, specificity, etc. ) is largely invariant \cite{Givon1983} even when the grammar expressing these functions may  change \cite{Komen2014}. Once again, the absence of statistical difference between child and adult TPRs indicates adult-like command of determiners with respect to discourse functions.

\section{Methodology}
\label{sec:cac}
Our main methodological contribution is a prompting scheme that turns child-caretaker  dialogue into controlled tests of linguistic knowledge in LLMs. For each target utterance \(U\) produced by children, we construct a discourse context \(C\) consisting of the preceding dialogue in the recording session, truncated only by the model's context window. Within \(U\), we identify the target position of a grammatical choice and elicit determiner preferences from the model\,---\,\emph{the} or \emph{a}\,---\,while preserving all surrounding contexts including the noun that the determiner combines with: Hence {\it Contextual Alternative Choice} (CAC) prompting (see Fig \ref{fig:cac-diagram}). 

Our empirical study is based on the Manchester Corpus  \cite{theakstonRolePerformanceLimitations2001} which contains regular recording sessions of 12 mother-child dyads aged 2-3, when determiners begin to be systematically used in speech.  Extrapolating from estimates of child-directed input \cite{Gilkerson2017}, these children will have heard approximately 10-20 million words in total. It is however more likely that the effective input is only 5-10 million words as the first year is primarily phonological development. These quantities provide useful checkpoints for LLMs as   models of child acquisition. 

The basic workflow of our study is illustrated in Fig \ref{fig:cac-diagram}. Both caretaker and child {\dn} pairs are extracted from the Manchester Corpus.  
The caretaker data is necessary for establishing the dialogue structure for the TPR analysis. Furthermore, because the caretaker's linguistic knowledge is not in doubt,  their data is also evaluated against the formal and functional benchmarks to provide a robustness test.  For each {\dn} pair in a child utterance \(U\), a language model undergoes CAC prompting with the context \(C\), defined above, to elicit determiner preferences in identical context as the child. This way, we obtain a model-generated corpus that parallels child-produced corpus: at each {\dn} site, the child has produced a particular determiner ({\it the} or {\it a}) while the model has produced a probabilistic distribution over them, which is then used to analytically derive its empirical predictions (Section \ref{subsec:evaluation}). 
Both child and model data are evaluated with our formal and functional benchmarks. 

Finally, we apply the benchmarks to the traditional forced-choice task by choosing determiners according to the model's maximum likelihood determiner (MLE) at each {\dn} site, in line with BLiMP and other such minimal pair tasks (Section 1).  To that end, we also report the extent to which a model's majority determiner choices agree with children, whom previous work has shown to be adult-like.  While we cannot exhaustively examine model output, strong divergence from child's usage may indict something amiss.

\subsection{Model inference}
\label{sec:model-scoring}
We evaluate 45 language models spanning three architecture classes: masked language models (MLM), autoregressive (AR) language models, and encoder-decoder (seq2seq) models. These are the  BabyLM baselines, winning submissions, and reference models publicly available as of March 2026. Full model metadata, including parameter counts, nominal training budgets, estimated total training exposure are reported in Appendix~\ref{app:models}, including whether the Manchester Corpus is known or likely to be included in training.

For each model, we apply CAC to obtain the logits of {\it the} and {\it a} at the site of each {\dn} combination in children's production, which are converted to a probabilistic distribution. Due to the differences in model architecture, there is some variation in model inference. For MLMs, we replace the target determiner with a single mask token and read off the logits at that position. Because all MLMs in our evaluation tokenize the determiners as single tokens, no multi-mask setup is required. Consistent with the child language studies, the phonological variant {\it an} is merged into {\it a} as a single choice. 

For AR models, we explicitly construct and score three full candidate utterances: one with \emph{the}, one with \emph{a}, and one with \emph{an} at the target position. Let \(U^{(\mathrm{the})}\), \(U^{(\mathrm{a})}\), and \(U^{(\mathrm{an})}\) denote these variants. Each candidate is scored by its conditional probability under the shared discourse context:
\[
p_\theta(U^{(k)} \mid C)
=
\prod_{t=1}^{T}
p_\theta(w_t^{(k)} \mid C, w_{<t}^{(k)}),
\]
where \(k \in \{\mathrm{the}, \mathrm{a}, \mathrm{an}\}\). Again we merge \(U^{(\mathrm{a})}\) and \(U^{(\mathrm{an})}\) as a single choice.

For seq2seq models with span-corruption objectives, we use the standard sentinel-token interface and score three teacher-forced target sequences corresponding to the same determiner alternatives. The encoder input replaces the target determiner with a sentinel token, and the decoder assigns 
\[
p_\theta(y^{(k)} \mid X)
=
\prod_{t=1}^{T'}
p_\theta(y_t^{(k)} \mid \mathrm{Enc}(X), y_{<t}^{(k)}),
\]
where \(X\) is the encoder input with the determiner replaced by the sentinel and \(k \in \{\mathrm{the}, \mathrm{a}, \mathrm{an}\}\). Finally, \emph{a} and \emph{an} are merged into a single choice.


\subsection{Model evaluation}
\label{subsec:evaluation}


Because the CAC prompting produces a probabilistic distribution of {\it the} and {\it a} for each {\dn} site in the model-generated corpus, it is possible to analytically derive the expected values of {\dn} overlap and TPRs. We discuss these in turn.

\paragraph{Formal \texttt{D}$\times$\texttt{N} productivity} For each child corpus which has \(N\) unique noun types and \(S\) {\dn} combinations, CAC prompts a model to generate a distribution over ({\it the}, {\it a}) at each {\dn} site.  Let \(k_n\) be the number of {\dn} sites for  a noun type \(n\), and $p_{n, i}$ be the probability of {\it the} at site $i$ where $i=1,2,\ldots k_n$. The expected value of overlap \(\widehat{O}\) for a model-generated corpus is the probability of a noun attested with both determiners averaged over all noun types:
\begin{equation}
    \widehat{O} = \frac{1}{N} \sum_{n \in N} \hat{o}_n, \,\,\hat{o}_n = 1 - \prod_{i=1}^{k_n} p_{n,i}
                  - \prod_{i=1}^{k_n} \bigl(1 - p_{n,i}\bigr) 
    \label{eq:analytical-overlap}
\end{equation}

The expected value of bias \(\hat{b}\) for the corpus can also be derived analytically. It is computed at the noun level (cf. Eq \ref{eq:bias}): for each noun type
\(n\), we sum the expected counts of \emph{the} and \emph{a} across all \(k_n\) sites and take the larger sum as that noun's contribution to the
preferred-determiner count:
\begin{equation}
    \hat{b} = \frac{1}{S} \sum_{n \in N}
    \max \left(
    \sum_{i=1}^{k_n} p_{n,i},
    \sum_{i=1}^{k_n} (1 - p_{n,i}) \right)
    \label{eq:analytical-bias}
\end{equation}
Together, \(N\), \(S\), and \(\hat{b}\) determine the predicted overlap under a
fully productive grammar (Eq. \ref{eq:app-mean-expected-overlap}; Appendix~\ref{app:dxn-derivation}). These values will be compared against the expected empirical  value \(\widehat{O}\); the absence of  significant difference supports  productivity. 

\paragraph{Determiner transitions.}
TPR calculation takes place over a pair of determiner transition $t = (C_t, M_t)$  sharing a topic noun,  with $C_t$ produced by the caretaker and $M_t$ generated by the model. Here, $C_t$ is either {\it the} or  {\it a} but the model choice $M_t$ is a probabilistic distribution ($M_t^{the}, M_{t}^{a}$). 
The expected value of TPR over all $T$ transitions is:
 \begin{equation}
     \widehat{\mathrm{TPR}} = \frac{1}{T} \sum_{t \in T} \bigl( M_t^{the} P(C_t = \mathrm{a}) + M_t^{a} P(C_t = \mathrm{the}) \bigr)
     \label{eq:analytical-tpr}
 \end{equation}
Model TPR is compared against children, caretakers, and a baseline from adult-to-adult conversations; see Section 4 for details. The comparison asks whether, in the statistical aggregate, a model uses determiners across conversational turns in a manner similar to human language users.

\section{Results}
\label{sec:results}
\begin{table*}[t]
\centering
\footnotesize
\caption{Human and representative model results. Means and SDs are computed over the 12 Manchester child dyads. Checkmarks indicate no significant deviation from benchmark (\(p > 0.05\)). Full results are in Table~\ref{tab:full-results}.}
\label{tab:summary-results}
\begin{adjustbox}{max width=\textwidth}
\begin{tabular}{llcccccc}
\toprule
Model & Arch. & Bias & Empirical overlap (SD) & Predicted overlap (SD) & TPR (SD) & DxN test & TPR test \\

\midrule
Children & Human & 0.834 & 0.251 (0.068) & 0.242 (0.105) & 0.226 (0.050) & {\bf $\checkmark$ (0.527)} & {\bf $\checkmark$ (0.484)} \\
Caretakers & Human & 0.815 & 0.301 (0.045) & 0.320 (0.070) & 0.200 (0.046) & {\bf $\checkmark$ (0.191)} & {\bf $\checkmark$ (0.256)} \\
\midrule
ltg-bert-bnc & MLM & 0.783 & 0.314 (0.068) & 0.287 (0.107) & 0.225 (0.048) & {\bf $\checkmark$ (0.138)} & {\bf $\checkmark$ (0.505)} \\
roberta-base & MLM & 0.776 & 0.323 (0.071) & 0.292 (0.112) & 0.239 (0.051) & {\bf $\checkmark$ (0.096)} & {\bf $\checkmark$ (0.142)} \\
t5-base & S2S & 0.753 & 0.347 (0.062) & 0.304 (0.102) & 0.239 (0.042) & $\times$ (0.030) & {\bf $\checkmark$ (0.085)} \\
babylm-baseline-100m-gpt-bert-causal-focus & AR & 0.782 & 0.314 (0.068) & 0.288 (0.106) & 0.247 (0.047) & {\bf $\checkmark$ (0.122)} & $\times$ (0.041) \\
gpt-bert-babylm-small & MLM & 0.902 & 0.185 (0.054) & 0.185 (0.091) & 0.431 (0.087) & {\bf $\checkmark$ (0.997)} & $\times$ (<0.001) \\
opt-125m & AR & 0.756 & 0.346 (0.067) & 0.303 (0.104) & 0.251 (0.044) & $\times$ (0.027) & $\times$ (0.018) \\
gpt2 & AR & 0.752 & 0.349 (0.069) & 0.305 (0.104) & 0.252 (0.042) & $\times$ (0.031) & $\times$ (0.012) \\
roberta-base-strict-2023 & MLM & 0.834 & 0.301 (0.067) & 0.249 (0.086) & 0.475 (0.062) & $\times$ (0.001) & $\times$ (<0.001) \\

\bottomrule
\end{tabular}
\end{adjustbox}
\end{table*}

We begin by testing human data against the benchmarks before turning to the models: both are summarized in Table \ref{tab:summary-results}.

\paragraph{Formal \texttt{D}$\times$\texttt{N} productivity.} In the Manchester corpus, children's mean empirical overlap was 0.251 (SD = 0.068) against the mean expected value of 0.242 (SD = 0.105),  and caretakers' mean empirical overlap was 0.301 (SD = 0.045) against the mean expected value of  0.320 (SD = 0.070). A paired \(t\)-test found no significant difference between empirical and expected values for either group (children: \(t(11)=0.653, p = 0.527\); caretakers: \(t(11)=-1.392, p = 0.191\)), replicating the findings of \citet{yangOntogenyPhylogenyLanguage2013} that determiners used by children (and adults) show productivity. Full dyad-level statistics are reported in Appendix~\ref{app:manchester-validation}.  

Note that the average bias value is remarkably similar between children (mean = 0.834, SD = 0.043) and caretakers (mean = 0.815, SD = 0.021): a paired \(t\)-test reveals no significant difference between children and caretakers (\(t(11) = 1.798, p = 0.100\)).  One-sample \(t\)-tests also found that neither group's mean bias differed significantly from the universal baseline value of 0.82 established from COCA \cite{COCA} (One sample t-test: children \(t(11) = 1.168, p = 0.268\); caretakers \(t(11) = -0.758, p = 0.464\)). This confirms the similarity and stability of  usage across English and in turn supports the robustness of the benchmark.

The caretakers' overlap values are considerably higher than the children's (paired \(t\)-test: t(11)=-3.616, p=0.004). Previous analyses \cite{Pine1996,Pine2013} regarded the difference as children's reduced productivity. However, the difference is due to adults talking more and having more opportunities for the nouns to be used with both determiners. This is reflected in the token/type ratio (\(S/N\)), which is higher for caretakers than for children (children: mean \(S/N = 4.192\); caretakers: mean \(S/N = 6.226\)) and is strongly correlated with empirical overlap in both groups (children: \(r(10)=.883, p<.001\); caretakers: \(r(10)=.695, p=.012\)).

\paragraph{Discourse-functional benchmark (TPR).} We first established the TPR baseline of 0.215 by examining  1,615 possible determiner transitions from adult-to-adult dialogue in CHILDES.  For the 12 children in the Manchester Corpus, the mean TPR when interacting with caretakers is 0.226 (SD = 0.050). For the 12 caretakers, the mean TPR when interacting with children is 0.200 (SD = 0.046), see Table \ref{tab:summary-results}. The children's TPR values do not significantly differ from their caretakers' on a paired \(t\)-test (\(t=2.061, p=0.064\)) even though caretakers sometimes repeat child speech verbatim.  Neither children nor caretakers show difference from the adult-to-adult baseline (One sample t-test: children \(t = 0.724\), \(p = 0.484\); caretakers \(t = 1.198\), \(p = 0.256\)). 
The statistical results suggest that the distribution of determiners in discourse is similar for children and adults alike, supporting earlier research that children use determiners appropriately.

\begin{figure*}[t]
    \centering
    \includegraphics[width=\textwidth]{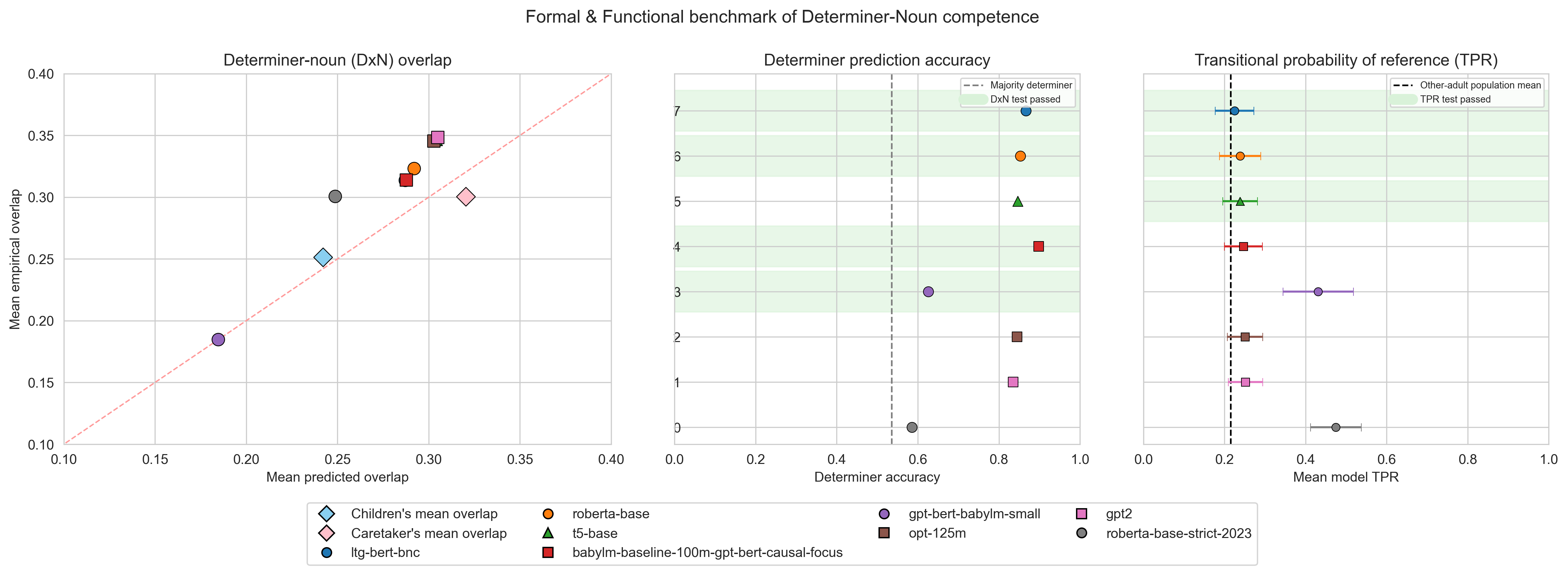}
    \caption{Formal and functional benchmarks for the representative models in Table~\ref{tab:summary-results}. Left: mean empirical vs.\ predicted \texttt{D}$\times$\texttt{N} overlap, with the dashed diagonal indicating identity and children/caretakers shown for reference. Middle: determiner prediction accuracy, with models passing the formal overlap test highlighted. Right: mean TPR across dyads, with the dashed line marking the adult population mean and models passing the TPR test highlighted.}
    \label{fig:main-benchmark}
\end{figure*}

\paragraph{LLM results.}
Table~\ref{tab:summary-results} summarizes representative model results compared to the human baselines across the two evaluation criteria, with the full results for all 45 models reported in Table~\ref{tab:full-results}. Figure~\ref{fig:main-benchmark} visualizes the representative models across all three diagnostic dimensions. 


Only two ({\tt ltg-bert-bnc} and {\tt roberta-base}) models passed both the formal overlap test and the discourse-functional TPR test. Both models, however, are trained on much larger amount of the data (100M and 30B words) than available to 2-3 year olds. The {\tt t5-base}  model, trained on even more data (156B words), also passes the TPR test but fails the overlap test. 

As shown in Table D1, 30 models satisfied the formal overlap test, including several models such as \texttt{ltgbert-10m-2024} and \texttt{gpt-bert-small} that were trained on an amount of data (10M words) comparable to that for 2-3 year olds. Setting aside four models whose outputs were degenerate due to always predicting the same determiner, this is an impressive result: Language models can learn the formal productivity of determiners and apply them interchangeably to nouns. However, none of these models pass the discourse-functional test, with their TPRs typically much higher:  0.433 for   {\tt ltgbert-10m-2024} and 0.438 for \texttt{gpt-bert-small}, compared to human speakers at 0.215. It is as if the models choose determiners stochastically with no regard for referential restrictions, as shown in the snippets of caretaker-child conversation.


Finally, we also evaluated the maximum likelihood determiner choices at each {\dn} site by the models. Doing so brings evaluation in line with classification tasks, and also allows us to measure
model accuracy in comparison to children. Several models achieved high accuracy: for example, \texttt{babylm-baseline-} \texttt{100m-gpt-bert-causal-focus} and \texttt{babylm-} \texttt{baseline-100m-gpt-bert-mixed} match child choices at 0.898 and 0.889, respectively, yet neither passed the TPR test, and only narrowly satisfied the overlap test.  This supports our earlier remark (Section 1) that   classification is not a rigorous demonstration of adult-like knowledge. 


\section{Related Work}
\label{sec:related-work}
{\bf Productivity test} The overlap measure has generated several quantitative assessments of child language. As noted earlier, they all require input data, which is available in very limited ways. For example, \citet{alhamaUsingComputationalModeling2024} mark determiner productivity as the recording session where  children use both determiners with at least two unique nouns. That, however, does not rule out the possibility of them having memorized them from prior but unrecorded input data. Similarly, \citet{meylanEmergenceAbstractGrammatical2017} estimate determiner probabilities for specific nouns, including the idiosyncratic {\it bathroom} and {\it bath}, to infer the degree of productivity.  Due to the sparsity of the input data, the inference model fails to converge for the majority of cases under study.  \citet{Valian2009} and \citet{Pine2013} compare children and caretakers directly and interpret a higher overlap value as greater productivity. But the overlap value itself can be misleading.  As \citet{yangOntogenyPhylogenyLanguage2013} observes, the overlap value of the Brown Corpus \cite{kuceraComputationalAnalysisPresentday1967}, a collection of print materials, is lower than some  children \cite{Pine1996} but professional writers clearly do not have a worse grammar than two-year-olds. 

\noindent {\bf CAC and LLM inference} 
CAC prompting adapts the cloze task \citep{Taylor1953}, which has previously served as a behavioral probe of language models across domains, from morphosyntactic analysis in context \citep{cotterell-etal-2018-conll, mccarthy-etal-2019-sigmorphon} to cross-modal brain decoding \citep{zou-etal-2022-cross}. CAC departs from prior cloze-style probes in two respects. First, it restricts the choice set to a closed grammatical alternation rather than  an open vocabulary, motivated by recent large-scale comparisons showing that LM next-token distributions and human cloze completions are not interchangeable \citep{jacobsLargescaleClozeEvaluation2024}.  Second, it is inspired by discourse cloze tasks  in  psycholinguistics  \citep{Chihara1977} and NLP \citep{Paperno2016}, rather than in the local sentence alone as in previous LLM work on determiners \citep{alhamaLinguisticProductivityCase2023, fiandraLargeLanguageModels2025}. It accentuates the recognition that language must accurately fulfill  communicative functions beyond the well-formedness of sentences.  

The probability distributions elicited by CAC are then treated as direct behavioral evidence, similar to a framework for LM surprisal analysis. Surprisal theory \citep{levyExpectationbasedSyntacticComprehension2008} establishes that the negative log-probability of a word in context is an interpretable signal of the model's own linguistic expectations, a link that appears to hold 
across languages \citep{wilcox-etal-2023-testing}. Model probabilities may thus provide a more reliable measure of linguistic knowledge than verbalized or prompted outputs even in instruction-tuned models \citep{Hu2023, kauf-etal-2024-log}. 

\section{Discussion and Conclusion}\label{sec:conclusion}
We introduce two quantitative benchmarks for language beyond classification that draw attention to the functional aspects of language that cannot be accurately assessed by well-formeness measures alone. Indeed, in the TPR analysis, both {\it the} and {\it a} are fully grammatical but the accurate choice must take discourse function and structure into account. We have shown that current language models trained on developmentally plausible data do not yet exhibit the knowledge of determiners that English-learning children demonstrate by age 2-3. Some, but but not all, models trained on much more data do meet the benchmarks but it is not clear what causes success or failure. The three-way dissociation between determiner prediction accuracy, formal \texttt{D}$\times$\texttt{N} productivity, and discourse-functional determiner use shows that these are different facets of competence, and that optimizing for one provides no guarantee of acquiring the others. 


Beyond the empirical findings, we also make a methodological contribution. The statistical benchmarks  require no access to training data and are applicable to any language sample produced by children or models alike. The overlap test in particular has been adapted to study  other linguistic phenomena in humans \citep{GMYang2017} and can be extended to LLMs as well. CAC, the prompting method introduced to elicit model output suitable for these benchmarks, is not specific to determiners: it can be applied to any grammatical choice at an attested site in naturalistic dialogue. 

More generally, our work calls for closer attention to child language research \citep{vazquezmartinezEvaluatingNeuralLanguage2023}. Early acquisition of determiners serves as a checkpoint for LLMs. Other aspects of morphology and grammar are learned later with frequent errors \citep{Ervin1964, Chomsky1969, yangPriceLinguisticProductivity2016} and can also serve as a constraining factor: LLMs may also  learn slowly with comparable errors. Finally, it is well known that the same element of grammar may be learned along different schedules across languages. For example, Italian- and Mandarin-learning children are adult-like in their usage of grammatical subjects around 2 \citep{Valian1991, Wang1992} but English-learning children only manage to reach adult-level by 4 \citep{Bloom1970}. Similarly, children acquiring English go through  years where a finite verb is substituted with an infinitive (e.g., {\it He eat cookies}) but children acquiring Spanish rarely do so \citep{LY2007}. And so on.  Such patterns of development, including variation within and across languages, have been amply documented and should shed light on the status of LLMs as cognitive models,  especially with the BabyLM project expanding the range of languages under investigation.
\newpage


\section*{Limitations}

\paragraph{Scope of the productivity test.}
The formal productivity test has additional constraints that limit its generality. First, the closed-class category must have exactly two members: the mathematical derivation of expected overlap is tractable for a binary category such as \{\textit{the}, \textit{a}\} but becomes intractable with additional members such as \textit{this} and \textit{that}. Second, the test assumes that the two categories combine in a fully interchangeable, statistically independent way. This characterizes determiners and nouns well, but not all grammatical combinations have this property. Not all transitive verbs passivize (\textit{John resembles Bill} cannot become \textit{*Bill was resembled by John}), not all dative verbs participate in both the double-object and \textit{to}-dative constructions, and so on. CAC can be deployed for such phenomena, but the formal productivity test cannot straightforwardly follow.

\paragraph{Corpus contamination.}
Many of the evaluated models were likely trained on corpora that include the Manchester data.  Despite our efforts to control for this using raw probability weights, contamination cannot be eliminated entirely, and results for models with known or likely Manchester exposure should be interpreted with appropriate caution.

\paragraph{Compute constraints.}
A single inference run through all 45 models required approximately 14 days of wall-clock time, which prevented evaluation of models at the 1B parameter scale and above. Larger models, and in particular frontier instruction-tuned and RLHF-trained systems, present an additional methodological consideration beyond scale: instruction tuning and reinforcement learning from human feedback optimize for conversational behavior that may interact with the CAC prompting interface in ways that differ from models trained on language modeling objectives alone. Whether such models meet the formal and functional criteria presented here is the topic of future work.





\appendix

\counterwithin{figure}{section}
\counterwithin{table}{section}
\renewcommand\thefigure{\thesection\arabic{figure}}
\renewcommand\thetable{\thesection\arabic{table}}

\newpage

\section{Derivation of the formal \texttt{D}$\times$\texttt{N} productivity test}
\label{app:dxn-derivation}

This appendix gives the derivation of the expected-overlap formula used in the main text for the formal \texttt{D}$\times$\texttt{N} productivity test. Following \citet{yangOntogenyPhylogenyLanguage2013}, the derivation assumes two statistical properties of the corpus: (i) singular noun frequencies follow an Zipf's Law, and (ii) nouns may exhibit an aggregate bias toward one determiner over the other.

\paragraph{Noun rank probabilities.} 
Let $S$ be the number of {\dn} combinations involving $N$ unique noun types. Assuming nouns follow Zipf's Law, the noun of rank \(r\) has probability
\begin{align}
    p_r &= \frac{1}{r^a H_{N}} \\
    \text{where}\quad
    H_{N} &= \sum_{i=1}^{N} \frac{1}{i}.
\end{align}

\paragraph{Determiner bias.} See Eq. \ref{eq:bias} in the main text. 

\paragraph{Expected overlap for a noun of rank \(r\).}
 For a noun of rank \(r\), the probability that it is never observed at all in \(S\) draws is
\[
(1-p_r)^S.
\]
Now consider the event that the noun is observed, but only with one determiner. Under the aggregate bias \(b\), the probability that a token is either not this noun or is this noun with its favored determiner is
\[
1-(1-b)p_r = b p_r + (1-p_r),
\]
so the probability that the noun never appears with its disfavored determiner is
\[
(b p_r + 1 - p_r)^S.
\]
Subtracting the probability that the noun is never observed at all gives the probability that it appears only with its favored determiner:
\[
(b p_r + 1 - p_r)^S - (1-p_r)^S.
\]
By the same reasoning, the probability that the noun appears only with its disfavored determiner is
\[
((1-b)p_r + 1 - p_r)^S - (1-p_r)^S.
\]

A noun contributes to overlap if it appears with both determiners at least once. Therefore, the expected overlap contribution of the noun of rank \(r\) is one minus the probabilities of the three complementary events: never observed, observed only with the favored determiner, and observed only with the disfavored determiner. This yields
\begin{equation}
\begin{aligned}
E_r
&=
1 - (1-p_r)^S \\
&\quad - \left[(b p_r + 1 - p_r)^S - (1-p_r)^S\right] \\
&\quad - \left[((1-b)p_r + 1 - p_r)^S - (1-p_r)^S\right].
\end{aligned}
\label{eq:app-expected-overlap}
\end{equation}

\paragraph{Expected overlap for the sample.}
The expected overlap for the full sample is the mean of these noun-level expectations over all noun ranks:
\begin{equation}
E[S]
=
\frac{1}{N}\sum_{r=1}^{N} E_r.
\label{eq:app-mean-expected-overlap}
\end{equation}

\paragraph{Empirical comparison.}
The empirical overlap statistic used in the main text is the proportion of noun types that occur with both determiners among all noun types that occur with either:
\begin{equation}
\mathrm{empirical}
=
\frac{1}{|N|}
\sum_{n\in N}
\mathbb{1}\left[\forall d\in\{\textit{the},\textit{a}\}, C_{d\times n}>0\right]\label{eq:app-empirical-overlap}
\end{equation}
The formal productivity test compares this empirical value with the expected value in Equation~\ref{eq:app-mean-expected-overlap}. If the two do not differ significantly, the sample is treated as consistent with a fully productive \texttt{D}$\times$\texttt{N} system.

\paragraph{Scope of the test.}
Although we apply the test here to English determiner--noun combinations, the logic is more general. As discussed by \citet{yangOntogenyPhylogenyLanguage2013} and \citet{goldin-meadowStatisticalEvidenceThat2017}, the same framework can be extended to other two-way combinatorial systems in which one category is small and closed, the other is larger and approximately Zipfian, and the question is whether the observed overlap is compatible with productive combination. \citet{yangOntogenyPhylogenyLanguage2013} also used it to reject productivty in Nim Chimpsky's ASL signs, providing independent support for distributional analysis of videos by his trainers \citep{Terrace1979}.

\section{Evaluated Models}
\label{app:models}

Tables~\ref{tab:models-mlm}--\ref{tab:models-s2s} list the language models evaluated in this study together with metadata relevant to the interpretation of our determiner experiments. Model names are given as \texttt{HuggingFace} repository paths where applicable. For each model, we report its parameter count, nominal training-data budget, the estimated total number of tokens seen during training, and whether the Manchester corpus is known or likely to be included in pretraining. We distinguish between nominal dataset size and total training exposure because several models were trained for multiple passes over relatively small corpora, so the total number of tokens seen during training exceeds the training budget allowance.

\paragraph{Column descriptions.}
\textbf{Dataset size} denotes the nominal size of the training corpus or benchmark track. \textbf{Total tokens seen} reports our compiled estimate of total training exposure. \textbf{Manchester in pretraining?} indicates whether the evaluation corpus is documented as included in pretraining, likely included, excluded, or unknown. We observed Manchester material directly in the BabyLM Strict and Strict-Small training corpora; accordingly, models trained on these datasets are marked as having included the Manchester corpus during pretraining.

\paragraph{Column descriptions.}
\textbf{Dataset size} denotes the nominal size of the training corpus or benchmark track. \textbf{Total tokens seen} reports our compiled estimate of total training exposure. \textbf{Manchester in pretraining?} indicates whether the Manchester corpus is known to be included in pretraining (\textit{yes}), confirmed absent (\textit{no}), likely but not independently confirmed (\textit{possibly}), or not determinable from available documentation (\textit{unknown}). Models trained on the BabyLM Strict (100M) or Strict-Small (10M) datasets are marked \textit{yes}: we confirmed that both tracks contain Manchester corpus material. Models trained on other corpora are marked \textit{yes} where independent confirmation was possible, \textit{possibly} where the training data composition makes inclusion likely but could not be verified, and \textit{unknown} where no information about corpus composition was available.

\begin{table*}[p]
\centering
\footnotesize
\caption{Masked language models evaluated in this study. Full Hugging Face repository paths are provided here; later tables use shortened model names for readability.}
\label{tab:models-mlm}
\begin{adjustbox}{max width=\textwidth}
\begin{tabular}{L{4.8cm}L{2.6cm}llll}
\toprule
Repository / family & Variant & Params. & Dataset size & Total tokens seen & Manchester? \\
\midrule
\multirow{3}{*}{\texttt{phueb/BabyBERTa}}
  & \texttt{-1} & 5M & 5M words & 50M words & No \\
  & \texttt{-2} & 5M & 5M words & 50M words & No \\
  & \texttt{-3} & 5M & 5M words & 50M words & No \\
\midrule
\multirow{2}{*}{\texttt{lgcharpe/ELC\_BERT}}
  & \texttt{\_small\_baby\_10M} & 24M & 10M words & \textasciitilde127B & Yes \\
  & \texttt{\_baby\_100M} & 98M & 100M words & \textasciitilde270B & Yes \\
\midrule
\multirow{4}{*}{\texttt{ltg/ltg-bert}}
  & \texttt{-bnc} & 98M & 100M words & 124B & No \\
  & \texttt{-babylm} & 98M & 100M words & 124B & Yes \\
  & \texttt{-10m-2024} & 30M & 10M words & Unknown & Possibly \\
  & \texttt{-100m-2024} & 119M & 100M words & Unknown & Yes \\
\midrule
\multirow{2}{*}{\shortstack[l]{\texttt{BabyLM-community/}\\\texttt{babylm-baseline-10m-gpt-bert}}}
  & \texttt{-masked-focus} & 31M & 10M words & 100M words & Yes \\
  & \texttt{-mixed} & 31M & 10M words & 100M words & Yes \\
\midrule
\multirow{2}{*}{\shortstack[l]{\texttt{BabyLM-community/}\\\texttt{babylm-baseline-100m-gpt-bert}}}
  & \texttt{-masked-focus} & 120M & 100M words & 1B words & Yes \\
  & \texttt{-mixed} & 120M & 100M words & 1B words & Yes \\
\midrule
\multirow{2}{*}{\texttt{roberta}}
  & \texttt{-base} & 125M & \textasciitilde30B & \textasciitilde2T & No \\
  & \texttt{-large} & 355M & \textasciitilde30--50B tokens & \textasciitilde2T & No \\
\midrule
\multirow{3}{*}{\texttt{nyu-mll/roberta-med-small}}
  & \texttt{-1M-1} & 45M & 1M words & \textasciitilde26B tokens & No \\
  & \texttt{-1M-2} & 45M & 1M words & \textasciitilde2.6B tokens & No \\
  & \texttt{-1M-3} & 45M & 1M words & \textasciitilde8.1B tokens & No \\
\midrule
\multirow{9}{*}{\texttt{nyu-mll/roberta-base}}
  & \texttt{-10M-1} & 125M & 10M words & \textasciitilde10B tokens & No \\
  & \texttt{-10M-2} & 125M & 10M words & \textasciitilde2.6B tokens & No \\
  & \texttt{-10M-3} & 125M & 10M words & \textasciitilde8.1B tokens & No \\
  & \texttt{-100M-1} & 125M & 100M words & \textasciitilde26B tokens & No \\
  & \texttt{-100M-2} & 125M & 100M words & \textasciitilde33B tokens & No \\
  & \texttt{-100M-3} & 125M & 100M words & \textasciitilde8.1B tokens & No \\
  & \texttt{-1B-1} & 125M & 1B words & \textasciitilde26B tokens & No \\
  & \texttt{-1B-2} & 125M & 1B words & \textasciitilde33B tokens & No \\
  & \texttt{-1B-3} & 125M & 1B words & \textasciitilde520B tokens & No \\
\midrule
\multirow{2}{*}{\texttt{babylm/roberta-base-strict}}
  & \texttt{-2023} & 125M & 100M words & 2B words & Yes \\
  & \texttt{-small-2023} & 125M & 10M words & 200M words & Yes \\\bottomrule
\end{tabular}
\end{adjustbox}
\end{table*}

\begin{table*}[p]
\centering
\footnotesize
\caption{Autoregressive models evaluated in this study. Full Hugging Face repository paths are provided here; later tables use shortened model names for readability.}
\label{tab:models-ar}
\begin{adjustbox}{max width=\textwidth}
\begin{tabular}{L{4.8cm}L{2.8cm}llll}
\toprule
Repository / family & Variant & Params. & Dataset size & Total tokens seen & Manchester? \\
\midrule
\multirow{2}{*}{\texttt{ltg/gpt-bert-babylm}}
  & \texttt{-small} & 30M & 10M words & 15.6B & Possibly \\
  & \texttt{-base} & 119M & 100M words & 31.25B & Possibly \\
\midrule
\multirow{2}{*}{\shortstack[l]{\texttt{BabyLM-community/}\\\texttt{babylm-baseline-10m-gpt-bert}}}
  & \texttt{-causal-focus} & 31M & 10M words & 100M words & Yes \\
  & \texttt{-mixed} & 31M & 10M words & 100M words & Yes \\
\midrule
\multirow{2}{*}{\shortstack[l]{\texttt{BabyLM-community/}\\\texttt{babylm-baseline-100m-gpt-bert}}}
  & \texttt{-causal-focus} & 120M & 100M words & 1B words & Yes \\
  & \texttt{-mixed} & 120M & 100M words & 1B words & Yes \\
\midrule
\texttt{timinar/baby-llama}
  & \texttt{-58m} & 58M & 10M words & Unknown & Yes \\
\midrule
\multirow{2}{*}{\texttt{babylm/babyllama}}
  & \texttt{-10m-2024} & 58M & 10M words & \textasciitilde100M words & Yes \\
  & \texttt{-100m-2024} & 58M & 100M words & \textasciitilde1B words & Yes \\
\midrule
\texttt{facebook/opt}
  & \texttt{-125m} & 125M & 180B & \textasciitilde180B tokens & Unknown \\
\midrule
\multirow{2}{*}{\texttt{babylm/opt-125m-strict}}
  & \texttt{-2023} & 125M & 100M words & 2B words & Yes \\
  & \texttt{-small-2023} & 125M & 10M words & 200M words & Yes \\
\midrule
\texttt{openai-community/gpt}
  & \texttt{2} & 124M & \textasciitilde8B tokens & \textasciitilde10B tokens & Unknown \\
\midrule
\multirow{2}{*}{\shortstack[l]{\texttt{BabyLM-community/}\\\texttt{babylm-baseline}}}
  & \texttt{-10m-gpt2} & 124M & 10M words & 100M words & Yes \\
  & \texttt{-100m-gpt2} & 124M & 100M words & 1B words & Yes \\
\bottomrule
\end{tabular}
\end{adjustbox}
\end{table*}

\begin{table*}[p]
\centering
\footnotesize
\caption{Encoder--decoder models evaluated in this study. Full Hugging Face repository paths are provided here; later tables use shortened model names for readability.}
\label{tab:models-s2s}
\begin{adjustbox}{max width=\textwidth}
\begin{tabular}{L{4.8cm}L{2.8cm}llll}
\toprule
Repository / family & Variant & Params. & Dataset size & Total tokens seen & Manchester? \\
\midrule
\texttt{google-t5/t5}
  & \texttt{-base} & 220M & 156B words & 34B tokens & Possibly \\
\midrule
\multirow{2}{*}{\texttt{babylm/t5-base-strict}}
  & \texttt{-2023} & 220M & 100M words & 2B words & Yes \\
  & \texttt{-small-2023} & 220M & 10M words & 200M words & Possibly \\
\bottomrule
\end{tabular}
\end{adjustbox}
\end{table*}

\section{Human validation on Manchester}
\label{app:manchester-validation}
As a validation of our preprocessing pipeline, we first applied the same extraction and overlap-computation procedure used throughout the paper to the human Manchester data. Singular noun phrases were identified automatically using the spaCy's \texttt{en-core-web-sm} noun chunk extractor \cite{honnibal-johnson-2015-improved}, after which we computed determiner--noun counts, empirical overlap, predicted overlap, and TPR for each child and caretaker sample. (Our results and conclusion hold when using the annotation provided by CHILDES.) We apply this same preprocessing pipeline to the language models evaluated in the main experiments.

The resulting raw overlap and TPR statistics are shown in Table~\ref{tab:manchester-validation}. These values closely reproduce the formal productivity pattern reported by \citet{yangOntogenyPhylogenyLanguage2013}: in the Manchester corpus, children's empirical \texttt{D}$\times$\texttt{N} overlap is statistically indistinguishable from their predicted overlap values (\(p=0.527\)). This result is consistent with the productivity expected under a fully productive system, and caretaker samples show the same general profile (\(p=0.191\)) despite differences in sample size and vocabulary. The same table also shows that both children and caretakers are statistically indistinguishable from the adult population TPR average (children: \(p=0.484\); caretakers: \(p=0.256\)). In this sense, our results align with the broader use of the overlap framework in \citet{goldin-meadowStatisticalEvidenceThat2017} and validate the TPR benchmark as well.

We include these statistics here for two reasons. First, they provide a sanity check on our implementation of both overlap analyses. Second, they corroborate our Manchester preprocessing pipeline as the basis for model-side evaluation, since the pipeline recovers the established human baseline on both the formal and functional benchmarks before being applied to BabyLM systems.
\begin{table*}[]
\centering
\small
\caption{Types (\(N\)), tokens (\(S\)), determiner bias score, token/type ratio (\(r\)), and observed (empirical) versus predicted overlap values for 12 children and their corresponding caretakers in the Manchester corpus. Singular noun phrases were extracted automatically using spaCy's noun chunker. At the group level, children's empirical overlap does not differ significantly from predicted overlap (\(p=0.527\)), and the same is true for caretakers (\(p=0.191\)), reproducing the formal productivity pattern reported in prior work and validating the preprocessing pipeline used throughout the paper.}
\label{tab:manchester-validation}
\begin{tabular}{llrrrrrrr}
\toprule
Dyad & Speaker & $|N|$ & $|S|$ & Bias & Empirical & Predicted & $n_\text{TPR}$ & TPR \\
\midrule
\multirow{2}{*}{Gail} & Child & 316 & 863 & 0.868 & 0.193 & 0.148 & 224 & 0.241 \\
                          & Caretaker & 838 & 3578 & 0.839 & 0.251 & 0.217 & 286 & 0.192 \\
\multirow{2}{*}{Dominic} & Child & 123 & 323 & 0.904 & 0.130 & 0.132 & 172 & 0.221 \\
                          & Caretaker & 539 & 4205 & 0.791 & 0.282 & 0.417 & 188 & 0.261 \\
\multirow{2}{*}{Becky} & Child & 364 & 1385 & 0.846 & 0.255 & 0.212 & 438 & 0.194 \\
                          & Caretaker & 592 & 3519 & 0.831 & 0.326 & 0.304 & 548 & 0.197 \\
\multirow{2}{*}{Liz} & Child & 312 & 1291 & 0.862 & 0.269 & 0.217 & 373 & 0.166 \\
                          & Caretaker & 619 & 3022 & 0.838 & 0.265 & 0.252 & 410 & 0.193 \\
\multirow{2}{*}{Carl} & Child & 407 & 3684 & 0.770 & 0.386 & 0.494 & 910 & 0.260 \\
                          & Caretaker & 516 & 3669 & 0.794 & 0.355 & 0.389 & 1077 & 0.208 \\
\multirow{2}{*}{Joel} & Child & 336 & 1020 & 0.889 & 0.179 & 0.144 & 331 & 0.145 \\
                          & Caretaker & 819 & 3550 & 0.836 & 0.220 & 0.223 & 445 & 0.133 \\
\multirow{2}{*}{Ruth} & Child & 203 & 747 & 0.798 & 0.246 & 0.258 & 279 & 0.272 \\
                          & Caretaker & 707 & 4668 & 0.796 & 0.310 & 0.355 & 372 & 0.183 \\
\multirow{2}{*}{Aran} & Child & 376 & 1635 & 0.813 & 0.258 & 0.263 & 802 & 0.261 \\
                          & Caretaker & 1072 & 8272 & 0.807 & 0.312 & 0.372 & 1079 & 0.152 \\
\multirow{2}{*}{Anne} & Child & 317 & 1170 & 0.815 & 0.290 & 0.233 & 500 & 0.168 \\
                          & Caretaker & 720 & 6083 & 0.831 & 0.368 & 0.386 & 657 & 0.158 \\
\multirow{2}{*}{John} & Child & 333 & 1615 & 0.807 & 0.279 & 0.296 & 473 & 0.307 \\
                          & Caretaker & 740 & 3876 & 0.789 & 0.341 & 0.301 & 526 & 0.276 \\
\multirow{2}{*}{Nicole} & Child & 195 & 492 & 0.860 & 0.210 & 0.152 & 234 & 0.214 \\
                          & Caretaker & 833 & 4372 & 0.836 & 0.262 & 0.260 & 280 & 0.179 \\
\multirow{2}{*}{Warren} & Child & 397 & 2314 & 0.782 & 0.320 & 0.355 & 826 & 0.262 \\
                          & Caretaker & 854 & 6080 & 0.797 & 0.316 & 0.367 & 944 & 0.265 \\
\bottomrule
\end{tabular}
\end{table*}

\begin{figure}[p]
    \centering
    \includegraphics[width=\columnwidth]{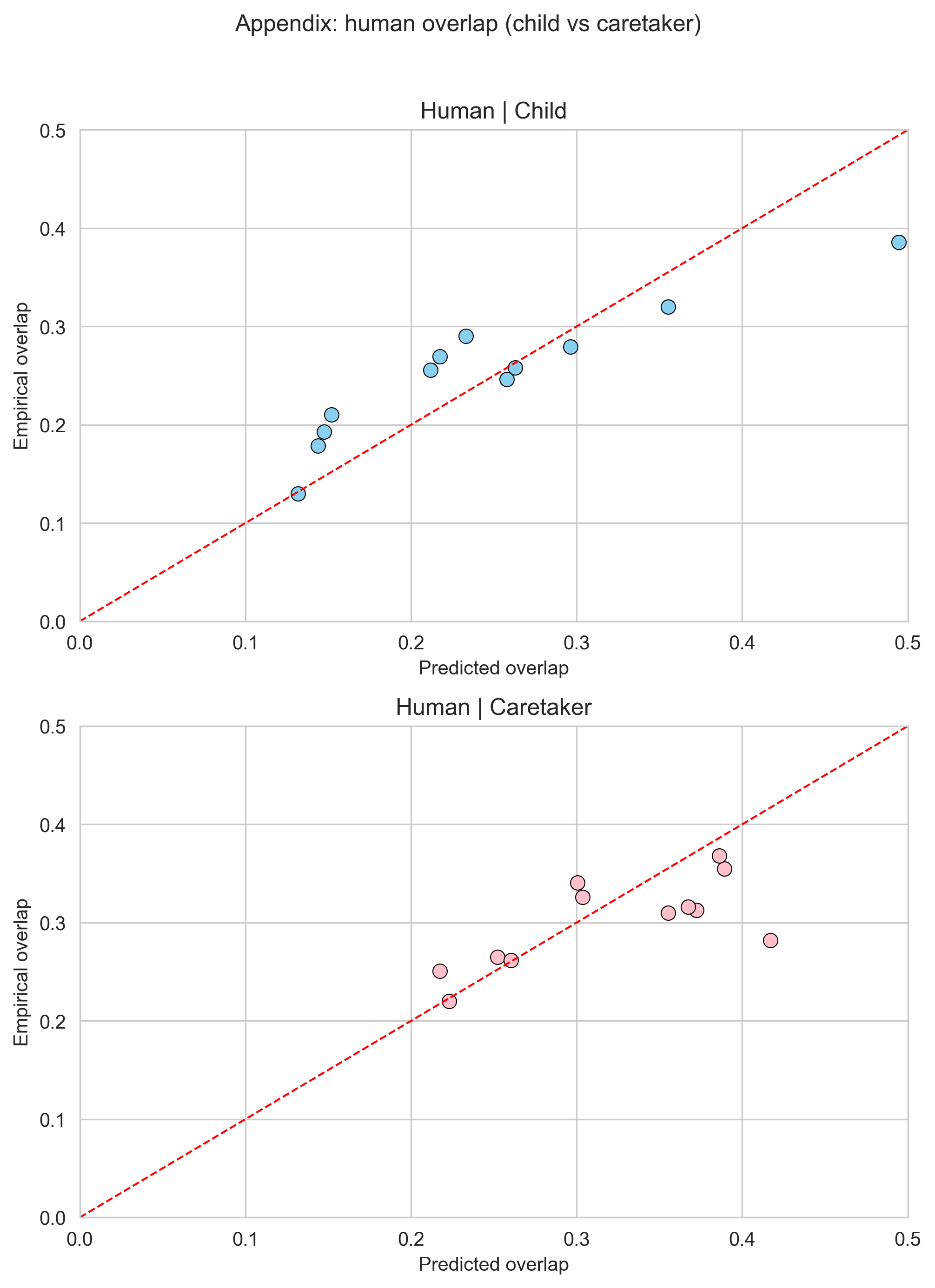}
    \caption{Full formal \texttt{D}$\times$\texttt{N} overlap results for 12 children (top) and their corresponding caretakers (bottom) in the Manchester corpus. No statistically significant difference is found between the empirical and predicted overlap scores (paired t-test $p=0.527$ children and $p=0.191$ for caretakers), meeting the formal \texttt{D}$\times$\texttt{N} overlap criterion.}
    \label{fig:appendix-overlap}
\end{figure}

\section{Full model results}
\label{app:full-results}

Table~\ref{tab:full-results} reports the full model-by-model results for all evaluated systems. It combines the formal \texttt{D}$\times$\texttt{N} productivity results and the functional discourse Transitional Probability of Reference (TPR) results into a single table.

For each model, we report mean empirical overlap and mean predicted overlap across the 12 Manchester child dyads, with standard deviations in parentheses, together with determiner bias, determiner accuracy, the outcome of the formal \texttt{D}$\times$\texttt{N} test, mean TPR, and the outcome of the TPR test. In the two test columns, the symbol indicates whether the model passed the corresponding criterion, and the exact \(p\)-value is given in parentheses in the same cell.

Models with non-significant differences between empirical and predicted overlap are treated as consistent with formal \texttt{D}$\times$\texttt{N} productivity. Models with non-significant differences under the TPR comparison are considered to have achieved discourse-appropriate functional usage of determiners. The main-text summary table is a compact subset of this full table.

Although four models with \texttt{bias}$\geq0.98$ appear to pass the formal \texttt{D}$\times$\texttt{N} test, this is an artifact of degenerate determiner predictions.  Concretely, all three \texttt{BabyBERTa} variants and \texttt{babylm-10m-gpt-bert-mixed (MLM)} overwhelmingly predict the same determiner ('a(n)'), as evidenced by the bias value being close to 1, yielding trivially small overlap values rather than substantive evidence of productivity.

\begin{table*}[]
\caption{Full model-by-model results for all evaluated systems. The table reports determiner bias, mean empirical and predicted \texttt{D}\(\times\)\texttt{N} overlap across the 12 Manchester child dyads, mean TPR, the outcomes of the formal \texttt{D}\(\times\)\texttt{N} and TPR tests, and determiner accuracy is the agreement with children when the MLE determiner is chosen by the model.  Checkmarks indicate passing (\(p > 0.05\)); exact \(p\)-values are shown in parentheses.}
\label{tab:full-results}
\begin{adjustbox}{max width=\textwidth}
\begin{tabular}{llccccccc}
\toprule
Model & Arch. & \texttt{bias} & \texttt{empirical} (SD) & \texttt{predicted} (SD) & TPR (SD) & \texttt{D$\times$N} test & TPR test & Accuracy \\
\midrule
Children & Human & 0.834 & 0.251 (0.068) & 0.242 (0.105) & 0.226 (0.050) & \textbf{$\checkmark$ (0.527)} & \textbf{$\checkmark$ (0.484)} & -- \\
Caretakers & Human & 0.815 & 0.301 (0.045) & 0.320 (0.070) & 0.200 (0.046) & \textbf{$\checkmark$ (0.191)} & \textbf{$\checkmark$ (0.256)} & -- \\
\midrule
BabyBERTa-1 & MLM & 0.996 & 0.014 (0.008) & 0.014 (0.007) & 0.497 (0.100) & \textbf{$\checkmark$ (0.057)} & $\times$ (<0.001) & 0.580 \\
BabyBERTa-2 & MLM & 0.999 & 0.005 (0.003) & 0.005 (0.003) & 0.497 (0.100) & \textbf{$\checkmark$ (0.445)} & $\times$ (<0.001) & 0.580 \\
BabyBERTa-3 & MLM & 0.999 & 0.006 (0.004) & 0.006 (0.004) & 0.497 (0.100) &\textbf{ $\checkmark$ (0.863) }& $\times$ (<0.001) & 0.580 \\
BabyLM-2026-Baseline-GPT2-Strict & AR & 0.701 & 0.381 (0.062) & 0.327 (0.096) & 0.393 (0.024) & $\times$ (0.007) & $\times$ (<0.001) & 0.653 \\
BabyLM-2026-Baseline-GPT2-Strict-Small & AR & 0.718 & 0.375 (0.064) & 0.321 (0.102) & 0.379 (0.036) & $\times$ (0.010) & $\times$ (<0.001) & 0.700 \\
ELC\_BERT\_baby\_100M & MLM & 0.799 & 0.296 (0.081) & 0.277 (0.118) & 0.299 (0.058) & \textbf{$\checkmark$ (0.308)} & $\times$ (<0.001) & 0.781 \\
ELC\_BERT\_small\_baby\_10M & MLM & 0.949 & 0.143 (0.042) & 0.116 (0.040) & 0.498 (0.090) & $\times$ (<0.001) & $\times$ (<0.001) & 0.580 \\
baby-llama-58m & AR & 0.741 & 0.352 (0.061) & 0.309 (0.100) & 0.271 (0.046) & $\times$ (0.025) & $\times$ (0.002) & 0.817 \\
babyllama-100m-2024 & AR & 0.775 & 0.319 (0.070) & 0.292 (0.108) & 0.256 (0.050) & \textbf{$\checkmark$ (0.126)} & $\times$ (0.017) & 0.885 \\
babyllama-10m-2024 & AR & 0.768 & 0.327 (0.067) & 0.296 (0.107) & 0.257 (0.048) & \textbf{$\checkmark$ (0.094)} & $\times$ (0.012) & 0.870 \\
babylm-baseline-100m-gpt-bert-causal-focus & AR & 0.782 & 0.314 (0.068) & 0.288 (0.106) & 0.247 (0.047) & \textbf{$\checkmark$ (0.122)} & $\times$ (0.041) & 0.898 \\
babylm-baseline-100m-gpt-bert-masked-focus & MLM & 0.980 & 0.062 (0.052) & 0.054 (0.041) & 0.489 (0.094) & $\times$ (0.043) & $\times$ (<0.001) & 0.582 \\
babylm-baseline-100m-gpt-bert-mixed & AR & 0.777 & 0.319 (0.069) & 0.291 (0.107) & 0.251 (0.046) & \textbf{$\checkmark$ (0.113)} & $\times$ (0.023) & 0.889 \\
babylm-baseline-100m-gpt-bert-mixed & MLM & 0.976 & 0.074 (0.038) & 0.067 (0.035) & 0.485 (0.095) & $\times$ (0.013) & $\times$ (<0.001) & 0.581 \\
babylm-baseline-100m-gpt2 & AR & 0.729 & 0.366 (0.075) & 0.317 (0.107) & 0.405 (0.041) & $\times$ (0.013) & $\times$ (<0.001) & 0.726 \\
babylm-baseline-10m-gpt-bert-causal-focus & AR & 0.762 & 0.334 (0.067) & 0.300 (0.109) & 0.267 (0.048) & \textbf{$\checkmark$ (0.072)} & $\times$ (0.003) & 0.861 \\
babylm-baseline-10m-gpt-bert-masked-focus & MLM & 0.957 & 0.118 (0.054) & 0.101 (0.046) & 0.488 (0.092) & $\times$ (0.003) & $\times$ (<0.001) & 0.581 \\
babylm-baseline-10m-gpt-bert-mixed & AR & 0.756 & 0.340 (0.069) & 0.303 (0.107) & 0.266 (0.043) & \textbf{$\checkmark$ (0.060)} & $\times$ (0.002) & 0.856 \\
babylm-baseline-10m-gpt-bert-mixed & MLM & 0.984 & 0.052 (0.035) & 0.049 (0.031) & 0.487 (0.096) & \textbf{$\checkmark$ (0.128)} & $\times$ (<0.001) & 0.582 \\
babylm-baseline-10m-gpt2 & AR & 0.740 & 0.348 (0.065) & 0.311 (0.108) & 0.389 (0.047) & \textbf{$\checkmark$ (0.066)} & $\times$ (<0.001) & 0.725 \\
gpt-bert-babylm-base & MLM & 0.916 & 0.175 (0.059) & 0.168 (0.081) & 0.441 (0.081) & \textbf{$\checkmark$ (0.560)} & $\times$ (<0.001) & 0.606 \\
gpt-bert-babylm-small & MLM & 0.902 & 0.185 (0.054) & 0.185 (0.091) & 0.431 (0.087) & \textbf{$\checkmark$ (0.997)} & $\times$ (<0.001) & 0.626 \\
gpt2 & AR & 0.752 & 0.349 (0.069) & 0.305 (0.104) & 0.252 (0.042) & $\times$ (0.031) & $\times$ (0.012) & 0.835 \\
ltg-bert-babylm & MLM & 0.797 & 0.297 (0.082) & 0.278 (0.119) & 0.314 (0.059) & \textbf{$\checkmark$ (0.289)} & $\times$ (<0.001) & 0.774 \\
ltg-bert-bnc & MLM & 0.783 & 0.314 (0.068) & 0.287 (0.107) & 0.225 (0.048) & \textbf{$\checkmark$ (0.138)} & \textbf{$\checkmark$ (0.505)} & 0.866 \\
ltgbert-100m-2024 & MLM & 0.739 & 0.352 (0.071) & 0.311 (0.110) & 0.308 (0.057) & \textbf{$\checkmark$ (0.055)} & $\times$ (<0.001) & 0.776 \\
ltgbert-10m-2024 & MLM & 0.792 & 0.287 (0.055) & 0.280 (0.106) & 0.433 (0.059) & \textbf{$\checkmark$ (0.714)} & $\times$ (<0.001) & 0.640 \\
opt-125m & AR & 0.756 & 0.346 (0.067) & 0.303 (0.104) & 0.251 (0.044) & $\times$ (0.027) & $\times$ (0.018) & 0.845 \\
opt-125m-strict-2023 & AR & 0.812 & 0.299 (0.060) & 0.263 (0.071) & 0.430 (0.068) & $\times$ (0.001) & $\times$ (<0.001) & 0.500 \\
opt-125m-strict-small-2023 & AR & 0.745 & 0.354 (0.067) & 0.308 (0.098) & 0.368 (0.030) & $\times$ (0.014) & $\times$ (<0.001) & 0.697 \\
roberta-base & MLM & 0.776 & 0.323 (0.071) & 0.292 (0.112) & 0.239 (0.051) & \textbf{$\checkmark$ (0.096)} & \textbf{$\checkmark$ (0.142)} & 0.853 \\
roberta-base-100M-1 & MLM & 0.756 & 0.343 (0.072) & 0.303 (0.115) & 0.267 (0.050) & \textbf{$\checkmark$ (0.052)} & $\times$ (0.004) & 0.787 \\
roberta-base-100M-2 & MLM & 0.752 & 0.343 (0.076) & 0.305 (0.117) & 0.285 (0.053) & \textbf{$\checkmark$ (0.070)} & $\times$ (<0.001) & 0.796 \\
roberta-base-100M-3 & MLM & 0.748 & 0.353 (0.076) & 0.308 (0.115) & 0.282 (0.051) & $\times$ (0.030) & $\times$ (<0.001) & 0.780 \\
roberta-base-10M-1 & MLM & 0.862 & 0.225 (0.076) & 0.227 (0.113) & 0.367 (0.068) & \textbf{$\checkmark$ (0.907)} & $\times$ (<0.001) & 0.702 \\
roberta-base-10M-2 & MLM & 0.823 & 0.278 (0.082) & 0.260 (0.120) & 0.338 (0.055) & \textbf{$\checkmark$ (0.328)} & $\times$ (<0.001) & 0.746 \\
roberta-base-10M-3 & MLM & 0.836 & 0.267 (0.081) & 0.250 (0.111) & 0.347 (0.065) & \textbf{$\checkmark$ (0.296)} & $\times$ (<0.001) & 0.729 \\
roberta-base-1B-1 & MLM & 0.770 & 0.331 (0.077) & 0.295 (0.119) & 0.278 (0.055) & \textbf{$\checkmark$ (0.072)} & $\times$ (0.002) & 0.786 \\
roberta-base-1B-2 & MLM & 0.747 & 0.351 (0.076) & 0.308 (0.115) & 0.283 (0.048) & $\times$ (0.031) & $\times$ (<0.001) & 0.797 \\
roberta-base-1B-3 & MLM & 0.775 & 0.324 (0.076) & 0.292 (0.117) & 0.285 (0.059) & \textbf{$\checkmark$ (0.101)} & $\times$ (0.002) & 0.805 \\
roberta-base-strict-2023 & MLM & 0.834 & 0.301 (0.067) & 0.249 (0.086) & 0.475 (0.062) & $\times$ (0.001) & $\times$ (<0.001) & 0.585 \\
roberta-base-strict-small-2023 & MLM & 0.777 & 0.355 (0.074) & 0.290 (0.102) & 0.487 (0.056) & $\times$ (0.002) & $\times$ (<0.001) & 0.584 \\
roberta-large & MLM & 0.766 & 0.331 (0.071) & 0.297 (0.116) & 0.270 (0.054) & \textbf{$\checkmark$ (0.111) }& $\times$ (0.005) & 0.832 \\
roberta-med-small-1M-1 & MLM & 0.754 & 0.331 (0.059) & 0.303 (0.099) & 0.475 (0.042) & \textbf{$\checkmark$ (0.149)} & $\times$ (<0.001) & 0.601 \\
roberta-med-small-1M-2 & MLM & 0.693 & 0.390 (0.070) & 0.332 (0.107) & 0.459 (0.019) & $\times$ (0.010) & $\times$ (<0.001) & 0.587 \\
roberta-med-small-1M-3 & MLM & 0.787 & 0.300 (0.061) & 0.284 (0.100) & 0.476 (0.051) & \textbf{$\checkmark$ (0.363)} & $\times$ (<0.001) & 0.594 \\
t5-base & S2S & 0.753 & 0.347 (0.062) & 0.304 (0.102) & 0.239 (0.042) & $\times$ (0.030) & \textbf{$\checkmark$ (0.085)} & 0.846 \\
t5-base-strict-2023 & S2S & 0.786 & 0.315 (0.083) & 0.286 (0.114) & 0.360 (0.067) & \textbf{$\checkmark$ (0.106)} & $\times$ (<0.001) & 0.748 \\
t5-base-strict-small-2023 & S2S & 0.657 & 0.420 (0.067) & 0.341 (0.090) & 0.483 (0.037) & $\times$ (<0.001) & $\times$ (<0.001) & 0.436 \\
\bottomrule
\end{tabular}
\end{adjustbox}
\end{table*}

\section{Full results figures}
\label{app:figures}

\begin{figure*}[p]
    \centering
    \includegraphics[height=0.9\textheight]{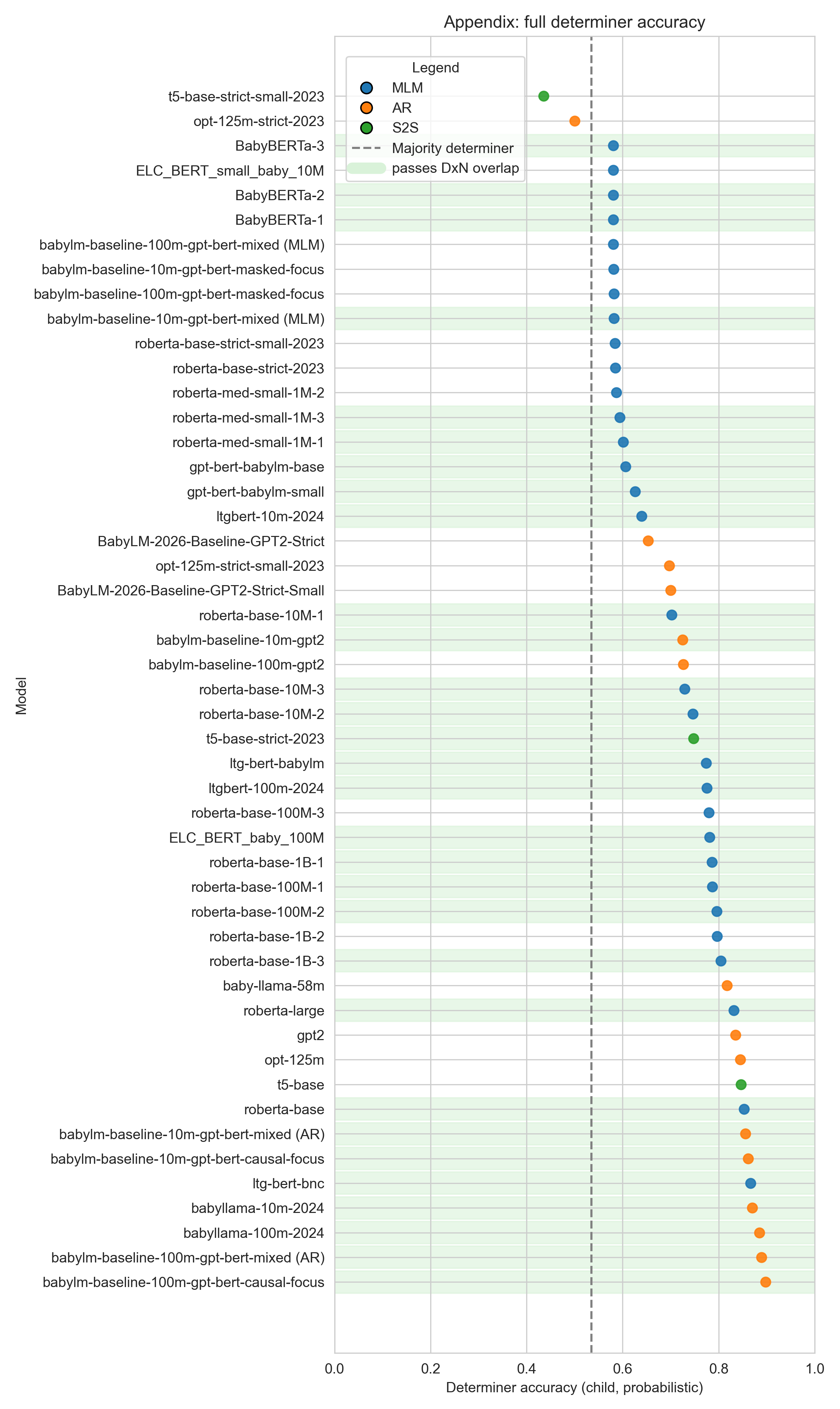}
    \caption{Full determiner prediction accuracy results for all evaluated models. The dashed vertical line indicates the accuracy of a majority-determiner baseline. Models that pass the formal \texttt{D}$\times$\texttt{N} overlap test are highlighted in green.}
    \label{fig:appendix-accuracy}
\end{figure*}

\begin{figure*}[p]
    \centering
    \includegraphics[height=0.9\textheight]{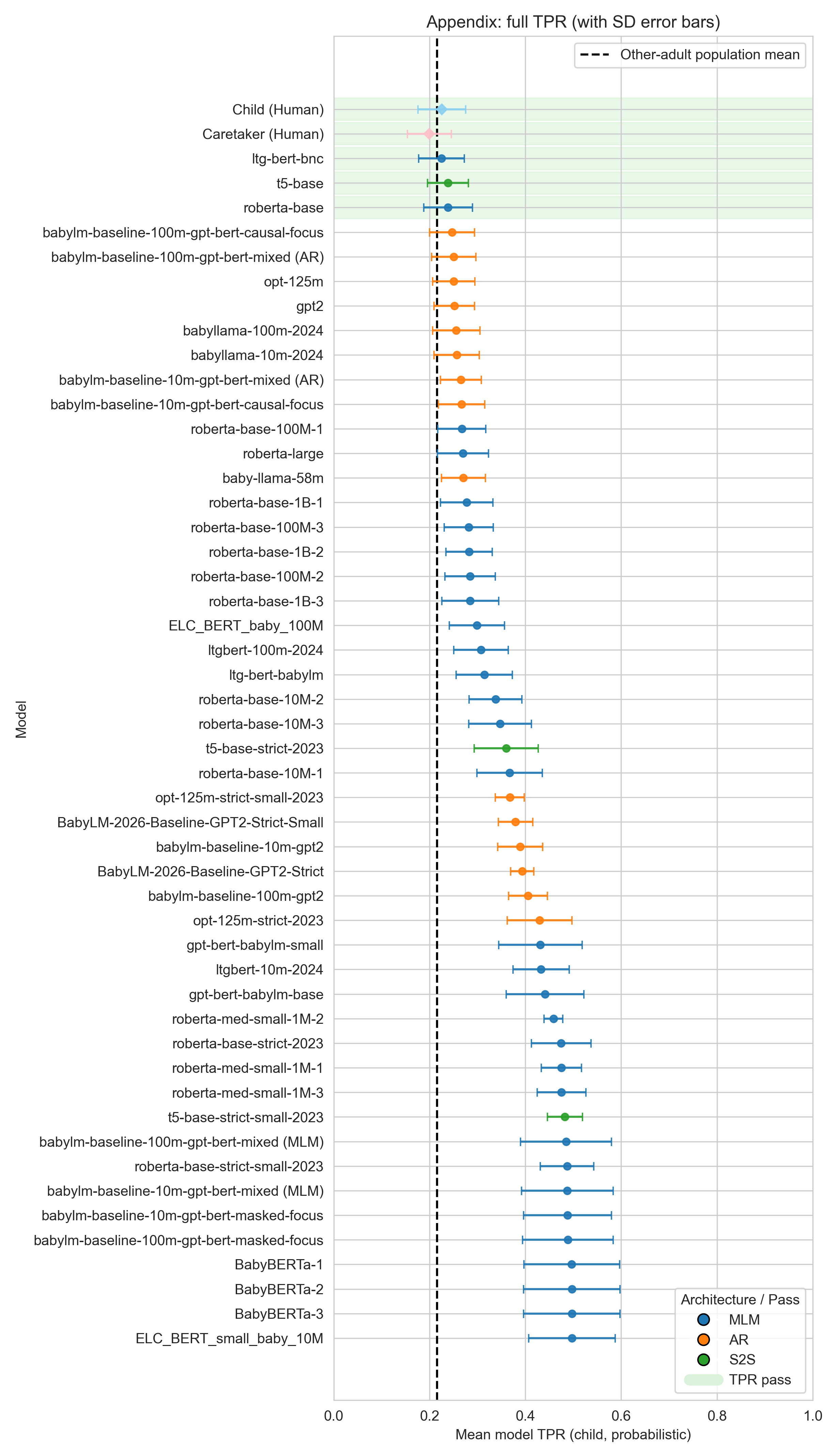}
    \caption{Full Transitional Probability of Reference (TPR) results for all evaluated models, with the 12 Manchester corpus children and corresponding caretakers at the top for reference. Points show mean TPR and whiskers show standard deviations across 12 dyads. The dashed vertical line indicates the adult population mean TPR, and models that pass the TPR test are highlighted in green.}
    \label{fig:appendix-tpr}
\end{figure*}

Figure~\ref{fig:appendix-overlap} visualizes the formal \texttt{D}$\times$\texttt{N} overlap results for the full set of evaluated models, with children and caretakers shown as human reference points. Figure~\ref{fig:appendix-accuracy} shows determiner prediction accuracy, with models that pass the formal overlap test highlighted. Figure~\ref{fig:appendix-tpr} shows mean TPR by model with standard-deviation whiskers, alongside the human child benchmark and models that pass the TPR criterion.

\begin{figure*}[p]
    \centering
    \includegraphics[width=\textwidth]{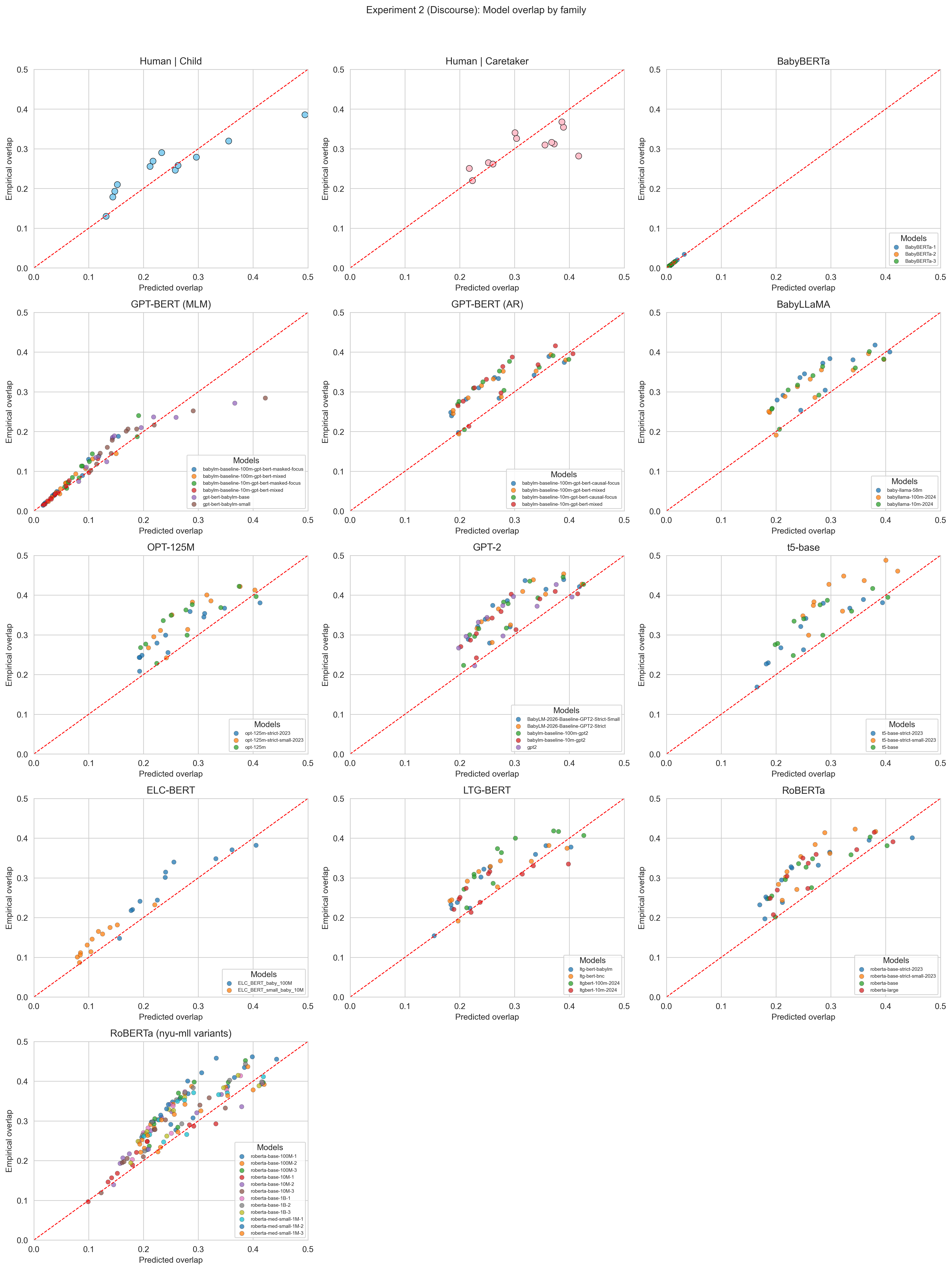}
    \caption{Full formal \texttt{D}$\times$\texttt{N} overlap results for all evaluated models. Each panel shows empirical overlap against predicted overlap for a model family, with the diagonal line indicating equality between observed and predicted overlap. The child and caretaker Manchester dyads are included as human reference panels.}
    \label{fig:appendix-overlap}
\end{figure*}

\end{document}